\documentclass[10pt,twocolumn,letterpaper]{article}

\usepackage{cvpr}
\usepackage{times}
\usepackage{epsfig}
\usepackage{graphicx}
\usepackage{amsmath}
\usepackage{amssymb}

\usepackage{multirow}
\usepackage{array,graphicx,subfigure}
\usepackage{bm}
\usepackage{bbm}
\usepackage{diagbox}
\usepackage{algorithm}
\usepackage{algpseudocode}

\usepackage[pagebackref=true,breaklinks=true,letterpaper=true,colorlinks,bookmarks=false]{hyperref}

\cvprfinalcopy 

\def\cvprPaperID{2763} 

\begin{document}

\title{SketchMate: Deep Hashing for Million-Scale Human Sketch Retrieval}

\author{Peng Xu$^{1,2}$~~~~Yongye Huang$^1$~~~~Tongtong Yuan$^1$~~~~Kaiyue Pang$^2$~~~~Yi-Zhe Song$^2$~~~~Tao Xiang$^2$\\
Timothy M. Hospedales$^3$~~~~Zhanyu Ma$^1$\thanks{Corresponding author.}~~~~Jun Guo$^1$\\
$^1$Beijing University of Posts and Telecommunications, China\\
$^2$SketchX, Queen Mary University of London, UK
$^3$The University of Edinburgh, UK\\
{\tt\small \{peng.xu, yongye, yuantt, mazhanyu, guojun\}@bupt.edu.cn}\\
{\tt\small \{p.xu, kaiyue.pang, yizhe.song, t.xiang\}@qmul.ac.uk}~~
{\tt\small t.hospedales@ed.ac.uk}
}

\maketitle

\begin{abstract}
We propose a deep hashing framework for sketch retrieval that, for the first time, works on a multi-million scale human sketch dataset. Leveraging on this large dataset, we explore a few sketch-specific traits that were otherwise under-studied in prior literature. Instead of following the conventional sketch recognition task, we introduce the novel problem of sketch hashing retrieval which is not only more challenging, but also offers a better testbed for large-scale sketch analysis, since: (i) more fine-grained sketch feature learning is required to accommodate the large variations in style and abstraction, and (ii) a compact binary code needs to be learned at the same time to enable efficient retrieval. Key to our network design is the embedding of unique characteristics of human sketch, where (i) a two-branch CNN-RNN architecture is adapted to explore the temporal ordering of strokes, and (ii) a novel hashing loss is specifically designed to accommodate both the temporal and abstract traits of sketches. By working with a 3.8M sketch dataset, we show that state-of-the-art hashing models specifically engineered for static images fail to perform well on temporal sketch data. Our network on the other hand not only offers the best retrieval performance on various code sizes, \textcolor{black}{but also yields the best generalization performance under a zero-shot setting and when re-purposed for sketch recognition. Such superior performances effectively demonstrate the benefit of our sketch-specific design.}

\end{abstract}

\section{Introduction}

Sketches are different to photos. They exhibit a high-level of abstraction yet are surprisingly illustrative. With just a few strokes, they are able to encode an appropriate level of semanticness that depicts objects and communicate stories (\eg, ancient cave drawings). Such unique characteristics of sketches, together with the prevalence of touchscreen devices, to a large extent drove the recent surge of sketch research. Problems studied so far range from sketch recognition~\cite{eitz2012humans,li2015freehand,yu2017sketch}, sketch-based image retrieval (SBIR)~\cite{yu2016sketch,sangkloy2016sketchy,xupeng2016Cross,xu2016instance,songjifei2016deep,xu2017cross,songjifei2017fine,li2017synergistic}, to sketch synthesis~\cite{li2017freesynthesis}.

Despite great strides made, a major obstacle facing all sketch research is the lack of freely available sketch data. Compared with photos where million-scale datasets had been readily accessible for almost a decade (\eg, ImageNet \cite{deng2009imagenet}), all aforementioned research worked with sub-million level crowd-sourced sketch datasets (20k for TU-Berlin \cite{eitz2012humans} and 75k for Sketchy~\cite{sangkloy2016sketchy}). These datasets served as key enablers for the community, though have very recently started to bottleneck the progress of sketch research -- sketch recognition performance had already gone far beyond human-level \cite{yu2017sketch} on TU-Berlin \cite{eitz2012humans}, and steadily approaching human performance~\cite{pang2017sketch} for the problem of SBIR on Sketchy ~\cite{sangkloy2016sketchy}.

In particular, two unique traits of human sketches had been mostly overlooked: (i) sketches are highly abstract and iconic, whereas photos are pixel perfect depictions, (ii) sketching is a dynamic process other than a mere collection of static pixels. Such oversights can be partially attributed to the lack of a large and diverse dataset of stroke-level human sketches, since more data samples are required to broadly capture (i) the substantial variances on visual abstraction, and (ii) the highly complex temporal stroke configurations -- an apple might look like an apple once drawn (though more abstract than photos), there is more than one way of drawing it. The seminal work of \cite{yu2017sketch} on sketch recognition tackled these problems to some extent yet were limited in that (i) sketches are treated as static pixelmaps, where deep architecture for feature learning is limited to variants of photo CNNs, and (ii) temporal ordering information is modeled coarsely by temporally segmenting one sketch into three separate pixelmaps, which are then encoded using a multi-branch CNN. The very recent work of \cite{quickdraw2017} was the first to fully acknowledge the temporal nature of sketches, and proposed a RNN-based generative model to synthesize novel sketches from scratch. In this paper, we combine RNN stroke modeling with conventional CNN under a dual-branch setting to learn better sketch feature representations. However, the problem of visual abstraction, especially how it can be accommodated under a large-scale retrieval setting remains unsolved.

In this paper, for the first time, we leverage on a newly released multi-million human sketch dataset \cite{quickdraw2017}, and introduce the novel problem of sketch hashing retrieval (SHR). Different to the conventional task of sketch recognition where classification is usually performed by computing feature distances in Euclidean space \cite{yu2017sketch}, given a query sketch, SHR aims to compute an exhaustive ranking of all sketches in a very large test gallery. It is thus a more difficult problem than sketch recognition, since (i) more discriminative feature representations are needed to accommodate the much larger variations on style and abstraction, and meanwhile (ii) a compact binary code needs to be learned to facilitate efficient large-scale retrieval. Importantly, the availability of such a large dataset enables us to better explore the aforementioned sketch-specific traits of being highly abstract and sequential in nature. In particular, we fully examine the temporal ordering of strokes through a two-branch CNN-RNN network, and address the abstraction problem by proposing a novel hashing loss that enforces more compact feature clusters for each sketch category in Hamming space.

\begin{figure}[!t]
\begin{center}
\includegraphics[width=0.45\textwidth]{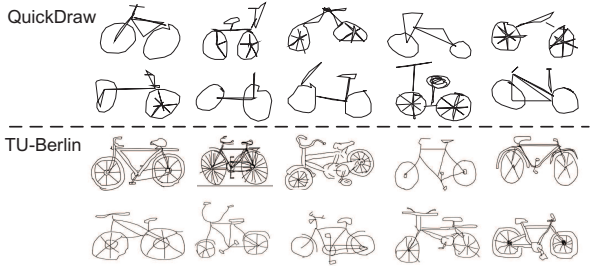}
\end{center}
   \caption{Sample sketches from QuickDraw-3.8M and TU-Berlin.}
\label{fig:dataset}
\end{figure}

More specifically, we first construct a dataset of 3,829,500 human sketches, by randomly sampling from every category of the Google QuickDraw dataset \cite{quickdraw2017}, which we term as ``QuickDraw-3.8M''. This dataset is highly noisy when compared with TU-Berlin, for that (i) users had only 20 seconds to draw, and (ii) no specific post-processing was performed. Figure \ref{fig:dataset} offers a visual comparison between the two datasets. We then analyze the intrinsic data traits of sketch and design a novel end-to-end deep hashing model to conduct fast retrieval.

The main contributions of this paper can be summarized as: (i) For the first time, we introduce the problem of sketch hashing retrieval on a multi-million scale human sketch dataset, and propose a deep hashing network that directly accommodates the key characteristics of human sketch. We show that our network is able to outperform state-of-the-art alternatives specifically designed for photo-photo and sketch-photo retrieval, highlighting the advantage of our sketch-specific design. Moreover, our network also achieves state-of-the-art performance when re-purposed for the task of sketch recognition, \textcolor{black}{and generalizes well under a zero-shot setting}. (ii) We propose a novel multi-branch CNN-RNN architecture that specifically encode\textcolor{black}{s} the temporal ordering information of sketches to learn a more fine-grained feature representation. We find that stroke-level temporal information is indeed helpful in sketch feature learning in that it alone can outperform CNN features for the sketch recognition task, and offers the best performance when combined with CNN features. (iii) We design a novel hashing loss to accommodate the abstract nature of sketches, especially on such a large dataset where noise is also present. More specifically, we propose a sketch center loss to learn more compact feature clusters for each object category and in turn improve retrieval performance.

The rest of the paper is organized as follows:
Section~\ref{sec:relatedwork} briefly summarizes related work.
Section~\ref{sec:metho} describes our proposed deep hashing model for large-scale sketch retrieval.
Experimental results and discussion are presented in Section~\ref{sec:experiments}.
Finally, we draw some conclusions in Section~\ref{sec:conclusion}.

\begin{figure*}[!t]
\begin{center}
\includegraphics[width=0.99\textwidth]{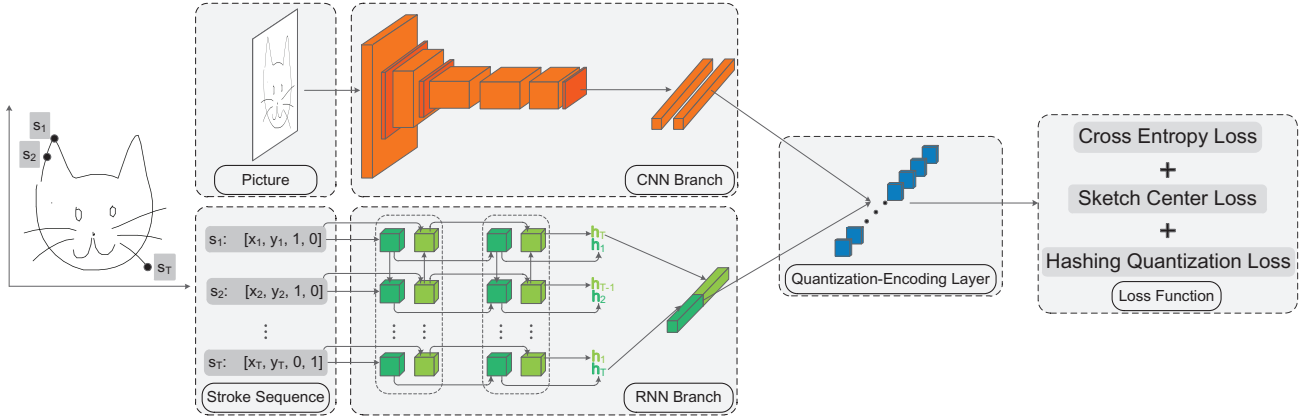}
\end{center}
   \caption{An illustration of our two-branch CNN-RNN deep sketch hashing retrieval network. Best viewed in color.}
\label{fig:pipeline}
\end{figure*}

\section{Related Work}
\label{sec:relatedwork}

\noindent\textbf{Sketch Dataset}\quad Collecting human sketches is naturally a cumbersome process -- they have to be drawn one by one other than crawled for free (as for photos). This largely contributed to the lack of large-scale human sketch datasets to date, especially those comparable to the scale of mainstream photo datasets \cite{deng2009imagenet}. Few medium-scale sketch datasets exist \cite{eitz2012humans, yu2016sketch, songjifei2017sketch, sangkloy2016sketchy}. They were mainly collected by resorting to crowd-sourcing platforms (\eg, Amazon Mechanical Turk) and asking the participant to either draw by hand or using a mouse. Albeit being large enough to train deep neutral networks, their sizes normally range from hundreds to thousands, thus inappropriate for large-scale deep hashing exploration that are inherently data-hungry. Very recently, this problem has been alleviated by Ha and Eck \cite{quickdraw2017}, who contributed a large-scale dataset containing $50$ millions of sketches crossing $345$ categories. These sketches are collected as part of a drawing game where participants has only 20 seconds to draw, hence are often very abstract and noisy. In this paper, we leverage on this dataset and study the novel problem of sketch hashing retrieval while proposing means of tackling the sketch-specific traits of abstraction and temporal ordering.

\noindent\textbf{Sketch Recognition}\quad A few shallow hand-crafted feature representations \cite{eitz2012humans,li2015freehand} have been proposed for sketch recognition. Albeit seeing some sketch-specific design, they are largely built from popular photo feature representations. The ground-breaking work of Yu \etal \cite{yusketch2015BMVC}, for the first time beats human performance on sketch recognition task by utilizing the discriminative power of a deep convolutional neural network. Subsequent work further exploited stroke-level temporal information by applying heuristic data augmentation \cite{yu2017sketch}. Our approach jointly explores static sketch visual characteristics and dynamic temporal sketching information in a single deep model. We show that it is superior to all existing models when re-purposed for the sketch recognition task.

\noindent\textbf{Deep Hashing Learning}\quad Hashing is an important research topic for fast image retrieval. Conventional hashing methods~\cite{andoni2006near,weiss2009spectral,gong2013iterative} mainly utilize hand-crafted features as image representations and propose various projections and quantization strategies to learn the hashing codes. Recently, deep hashing learning has shown superiority on better preserving the semantic information when compared with shallow methods~\cite{xia2014supervised,lin2015deep,shen2015supervised}. In the initial attempt, feature representation and hashing codes were learned in separate stages~\cite{xia2014supervised}, where subsequent work ~\cite{lin2015deep,zhao2015deepsemantic,liu2016deepsupervised} suggested superior practice through joint end-to-end training. To our best knowledge, only one previous work \cite{liu2017deepsketchhashing} has specifically designed a deep hashing framework targeted on sketch data. They introduced a semi-heterogeneous deep architecture by incorporating cross-view similarity and a cross-category semantic loss. Despite its superior performance, sketch specific traits such as stroke ordering and drawing abstraction were not accommodated for. The dataset \cite{sangkloy2016sketchy} they evaluated on is also arguable too small to truly show for the practical value of a deep hashing framework. We address these issues by working with a much larger human sketch dataset, and designing sketch-specific solutions that are crucial for million-scale retrieval.

\begin{figure*}[!t]
	\centering
	\subfigure[cross entropy loss]{
		\label{fig:softmax_fig}
		\includegraphics[width=0.25\textwidth]{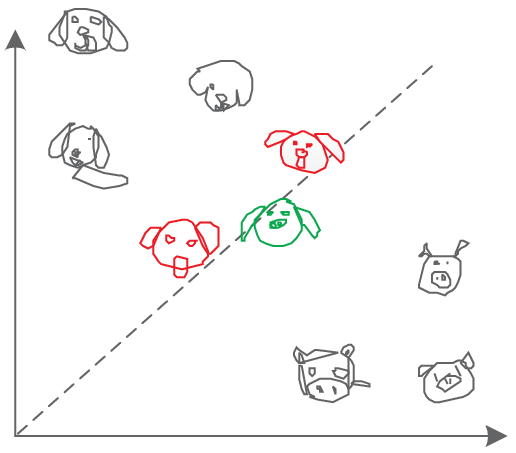}}
	\subfigure[cross entropy loss + common center loss]{
		\label{fig:softmax_centerloss_fig}
		\includegraphics[width=0.25\textwidth]{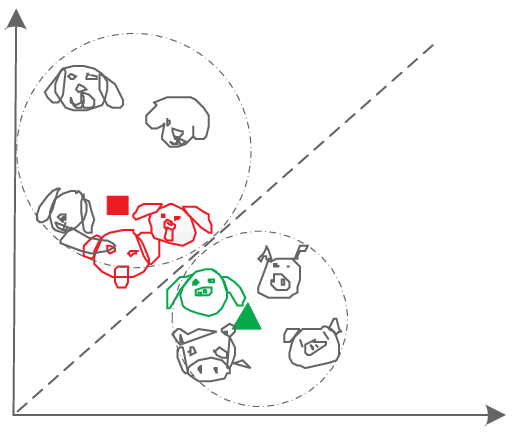}}
    \subfigure[cross entropy loss + sketch center loss]{
		\label{fig:softmax_sketchcenterloss_fig}
		\includegraphics[width=0.25\textwidth]{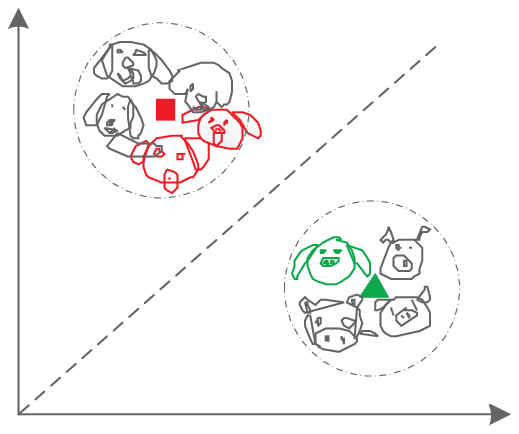}}
	\caption{Geometric interpretation of sketch feature layout obtained by different loss function. The dashed line denotes the softmax decision boundary. See details in text.}
	\label{fig:softmax_centerloss}
\end{figure*}

\section{Methodology}
\label{sec:metho}
\subsection{Problem Formulation}

Let $\mathcal{K} = \{{K}_n=({\bf P}_n,{\bf S}_n)\}_{n=1}^{N}$ be $N$ sketch sample pairs crossing $L$ possible categories
and $\mathcal{Y} = \{y_n\}_{n=1}^{N}$ be their respective category labels.
Each sketch sample ${K}_n$ consists of a sketch ${\bf P}_n$ in raster pixel space and a corresponding sketch segment sequence ${\bf S}_n$. We aim to learn a mapping $\mathcal{M}:\mathcal{K} \rightarrow \{0, 1\}^{D \times N}$,
which represents sketches as $D$-bit binary codes
${\bf B} = \{{\bf b}_n\}_{n=1}^{N} \in \{0, 1\}^{D \times N}$, while maintaining relevancy in accordance with the semantic and visual similarity.

\subsection{Two-branch CNN-RNN Network}
\noindent\textbf{Overview}\quad As previously stated, learning discriminative sketch features is a very challenging task due to the high degree of variations in style and abstraction. This problem is made worse under a large-scale retrieval setting since better feature representations are needed for more fine-grained feature comparison. Despite shown to be successful on a much smaller sketch dataset \cite{yu2017sketch}, CNN-based network completely abandons the inherent stroke-level temporal information of human sketches, which can now be modeled by a RNN network thanks to the seminal work by \cite{quickdraw2017}. In this work, we for the first time, propose to combine the best from the both world for human sketches -- utilizing CNN to extract abstract visual concepts and RNN to model human sketching temporal orders. With additional discriminative power (temporal cue) injected in, we expect this can lead to better feature learning.

\noindent\textbf{Two-branch Late-fusion}\quad As illustrated in Figure \ref{fig:pipeline}, our two-branch encoder consists of three sub-modules: (1) a CNN encoder takes in a raster pixel sketch and translates into a high-dimensional space; (2) a RNN encoder takes in a vector sketch and outputs its final time-step state; (3) branch interaction via a late-fusion layer by concatenation. This enables our learned feature to benefit from both vector and raster sketch.

\noindent\textbf{Quantization Encoding layer}\quad After the final fusion layer, we have to encode that deep feature into the low-dimensional real-valued hashing feature ${\bf f}_{n}$ (one fully connected layer with sigmoid activation), which will be further transformed to the hashing code, ${\bf b}_{n}$. The transformation function goes as follows:


\begin{equation}
\label{equ:quantization}
{\bf b}_n = sgn({\bf f}_n - {\bf 0.5}),~~n \in (1, N).
\end{equation}

\noindent\textbf{Learning Objective}\quad To obtain the hashing feature ${\bf f}_{n}$ and hashing code ${\bf b}_{n}$, we could train the network end-to-end using two common losses similar to those found in image hashing networks \cite{lin2015deep}. The first comes with the cross entropy loss (CEL) for $\mathcal{K}$ calculated on $L$-way softmax:
\begin{equation}
\label{equ:cel_loss}
\begin{split}
\mathcal{L}_{cel} = & \frac{1}{N} \sum_{n = 1}^{N} - \log \frac{\mathrm{e}^{{\bf W}_{y_n}^{T}{\bf f}_n+\widehat{b}_{y_n}}}{\sum_{j=1}^{L} \mathrm{e}^{{{\bf W}_j^{T}{\bf f}_n+\widehat{b}_j}}},
\end{split}
\end{equation}
where ${\bf W}_j \in \mathbb{R}^{D}$ is the $j$th column of the weights ${\bf W} \in \mathbb{R}^{D \times L}$ between the quantization-encoding layer and $L$-way softmax outputs. $\widehat{b}_j$ is the $j$th term of the bias $\widehat{{\bf b}} \in \mathbb{R}^L$. Quantization loss (QL) is used to reduce the error caused by quantization-encoding:
\begin{equation}
\label{equ:ql_loss}
\begin{split}
\mathcal{L}_{ql} = \frac{1}{N} \sum_{n = 1}^{N} \|{\bf b}_n - {\bf f}_n\|_2^2,~ s.t.~{\bf b}_n \in \{0, 1\}^{D}.
\end{split}
\end{equation}

\subsection{Sketch Center Loss}
\label{sec: scl}
In theory, these two losses should perform reasonably well on discriminating category-level semantics, however, our large-scale sketch dataset presents an unique challenges -- sketch are highly abstract, often making semantically different categories to exhibit similar appearance (see Figure \ref{fig:softmax_fig} for an example of `dog' vs. `pig'). We need to make sure such abstract nature of sketches do not hinder overall retrieval performance. The common center loss (CL) was proposed in \cite{wen2016discriminative} to tackle such a problem by introducing the concept of class center, ${\bf c}_{y_{n}}$, to characterize the intra-class variations. Class centers should be updated as deep features change, in other words, the entire training set should be taken into account and features of every class should be averaged in each iteration. This is clearly unrealistic and normally compromised by updating only within each mini-batch. This problem is even more salient under our sketch hashing retrieval setting -- (1) for million-scale hashing, updating common center within each mini-batch can be highly inaccurate and even misleading (as shown in later experiments), and this problem is worsened by the abstract nature of sketches in that only seeing sketches within one training batch doesn't necessarily provide useful and representative gradients for class centers; (2) despite of more compact internal category structures (Figure \ref{fig:softmax_centerloss_fig} with common center loss, there is no explicit constraint to set apart between each, as a direct comparison with Figure \ref{fig:softmax_sketchcenterloss_fig}.


These issues call for a sketch-specific center loss that is able to deal with million-scale hashing retrieval. For sketch hashing, we need compact and discriminative features to aggregate samples belonging to the same category and segregate the visually confusing categories. Thus, an natural intuition would be: is it possible if we can find a \textit{fixed} but representative center feature for each class, so to avoid the computational complexity during training, and meanwhile enforcing semantics between sketch categories.

We propose \textit{sketch center loss} that is specifically designed for million-scale sketch hashing retrieval. This is done by (i) first pretraining CNN-RNN separately for sketch recognition task and then fine-tuning with our full model, both with softmax cross entropy loss only; (ii) obtain class feature center $c_{y_{n}}$ by calculating the mean of the hashing feature $f_{n}$ for the \textit{noise-removal} sketches (detailed later) of that class based on the pretrained model. By doing so, in the final fine-tuning stage, we train end-to-end with a fixed center for each class, thus providing meaningful gradients during each training iteration, and we empirically find a significant performance boost under this sketch-specific center loss. We hence define our sketch center loss as:

\begin{equation}
\label{equ:sketch_center_loss}
\mathcal{L}_{scl} = \frac{1}{N} \sum_{n = 1}^{N} \|{\bf f}_n - {\bf c}_{y_n}\|_2^2~,
\end{equation}

\begin{figure}[t]
\begin{center}
\includegraphics[width=0.42\textwidth]{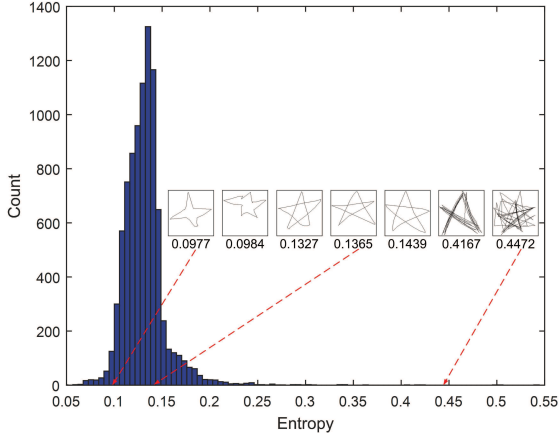}
\end{center}
   \caption{\textcolor{black}{Image entropy histogram of `stars' in our training set. The blue bars denote the bin counts within different entropy ranges. Some representative sketches corresponding to different entropy value ranges are illustrated. See details in text.}}
\label{fig:star}
\end{figure}

\noindent\textbf{Noise Removal with Image Entropy}\quad Key ingredient to a successful sketch center loss is the guarantee of non-noisy data (outliers), as it will significantly affect the class feature centers. However, sketch data collected with crowdsourcing are inevitable to noise, where we propose a noisy data removal technique to alleviate such issue by resorting to image entropy. Given a category of sketch, we can get entropy for each sketch and the overall entropy distribution on a category basis. We empirically find that keeping the middle $90\%$ of each category as normal samples gives us best results. In Figure \ref{fig:star}, we visualize the entropy histogram of star samples in our training set. If we choose the middle $90\%$ samples as normal samples for star category, we can calculate and get the $0.05$ and $0.95$ percentiles of star images entropy as $0.1051$ and $0.1721$, respectively. We then treat the remaining samples as outliers or noise points (entropy $\in~[0,0.1051)~\bigcup~(0.1721,1]$). It can be observed that low entropy sketches tend to be overly-abstract, yet high entropy ones being messy, sometimes with meaningless scribbles. Nevertheless, sketch data falling into middle entropy range present more consistent and reasonable drawings.

\noindent\textbf{Full Learning Objective}\quad By combining the above, our full objective becomes:
\begin{equation}
\label{equ:full}
\mathcal{L}_{full} = \mathcal{L}_{cel} +\lambda_{scl}\mathcal{L}_{scl} + \lambda_{ql}\mathcal{L}_{ql},
\end{equation}
where $\lambda_{scl}, \lambda_{ql}$ control the relative importance of each loss. The detailed training and optimization procedures are described in Algorithm~\ref{alg:1}.

\begin{algorithm}
    	\caption{Algorithm for the proposed deep sketch hashing model via multiple staged-pretraining.}
  	\label{alg:1}
    \noindent\textbf{Input:} $\mathcal{K} = \{{K}_n=({\bf P}_n,{\bf S}_n)\}_{n=1}^{N}$, $\mathcal{Y} = \{y_n\}_{n=1}^{N}$.
        \begin{algorithmic}[1]
        \State Train CNN from scratch using $\{{\bf P}_n\}_{n=1}^{N}, \mathcal{L}_{cel}$.
        \State Train RNN from scratch using $\{{\bf S}_n\}_{n=1}^{N}, \mathcal{L}_{cel}$.
        \State Parallelly connect pretrained RNN and CNN branches via hashing quantization-encoding layer. Fine-tune the fused model using $\mathcal{L}_{cel}$ without binary constraint.
        \State Calculate class feature centers basing on the pretrained model in step $3$. Fine-tune the whole network using $\mathcal{L}_{cel} + \lambda_{scl}\mathcal{L}_{scl}$.
        \State Finally, train the model subjected to our full learning objective, $\mathcal{L}_{full}$, with procedures as following ($t$ represents current iteration):
        \For{number of training iterations}
       		\For{a fixed number of iterations}
                \State Fix ${\bf b}_n^{t}$, update $\Theta_{cnn}, \Theta_{rnn}$ using~(\ref{equ:full}).
            \EndFor
            \State Fix $\Theta_{cnn}, \Theta_{rnn}$, calculate ${\bf b}_n^{t+1}$ using~(\ref{equ:quantization}).
        \EndFor
        \end{algorithmic}
        \noindent\textbf{Output:} Network parameters: $\Theta_{cnn}$~and~$\Theta_{rnn}$. Binary hash code matrix ${\bf B} \in \mathbb{R}^{D \times N}.$
    \end{algorithm}

\begin{table*}[!tbp]
\footnotesize
\begin{center}
\resizebox{0.9\textwidth}{!}{
\begin{tabular}{|c|l|c|c|c|c|c|c|c|c|}
\hline
\multirow{2}{*}{No.} & \multirow{2}{*}{Model} & \multicolumn{4}{c|}{Mean Average Precision} & \multicolumn{4}{c|}{Precision @200} \\
& & 16 bits & 24 bits & 32 bits & 64 bits & 16 bits & 24 bits & 32 bits & 64 bits \\
\hline\hline
1 & DLBHC~\cite{lin2015deep} & 0.5453 & 0.5910 & 0.6109 & 0.6241 &0.5142 &0.5917 &0.6169 &0.6403 \\
2 & DSH-Supervised~\cite{liu2016deepsupervised} & 0.0512 & 0.0498 & 0.0501 & 0.0531 &0.0510 &0.0512 &0.0501 &0.0454  \\
3 & DSH-Sketch~\cite{liu2017deepsketchhashing} & 0.3855 & 0.4459 & 0.4935 & 0.6065 &0.3486 &0.4329 &0.4823 &0.6040 \\
\hline
\hline
4 & Our+CEL & 0.5969 &0.6196 & 0.6412 & 0.6525 & 0.5817 & 0.6292 &0.6524 &0.6730     \\
5 & Our+CEL+CL & 0.5567 & 0.5856 & 0.5911 & 0.6136 & 0.5578 & 0.6038 &0.6140 &0.6412      \\
6 & Our+CEL+SCL & 0.6016 & 0.6371 & 0.6473 & 0.6767 & 0.5928 & 0.6298 & 0.6543&0.6875   \\
\hline
\hline
7 & Our+CEL+SCL+QL (Full) & $\bm{0.6064}$ & $\bm{0.6388}$ &$\bm{0.6521}$ & $\bm{0.6791}$ & \textbf{0.5978} & \textbf{0.6324} &\textbf{0.6603} &\textbf{0.6882}   \\
\hline
\end{tabular}}
\end{center}
\caption{Comparison with state-of-the-art deep hashing methods and our model variants on on QuickDraw-3.8M retrieval gallery.}
\label{table:compare_with_deep_baseline}
\end{table*}

\begin{table*}
\footnotesize
\begin{center}
\resizebox{0.9\textwidth}{!}{
\begin{tabular}{|c|c|c|c|c|c|c|c|c|c|}
\hline
 \multicolumn{2}{|c|}{~} & \multicolumn{6}{c|}{Unsupervised} & \multicolumn{2}{c|}{Supervised} \\
 \multicolumn{2}{|c|}{~} & PCA-ITQ~\cite{gong2013iterative} & LSH~\cite{andoni2006near} & SH~\cite{weiss2009spectral} & SKLSH~\cite{raginsky2009locality} & DSH~\cite{jin2014density} & PCAH~\cite{wang2006PCAH} & SDH~\cite{shen2015supervised} & CCA-ITQ~\cite{gong2013iterative} \\
\hline\hline
 \multirow{5}{*}{HOG} & 16 bits & 0.0222 & 0.0110 & 0.0166 & 0.0096 &	0.0186 & 0.0166 & 0.0160 & 0.0185 \\
                      & 24 bits & 0.0237 & 0.0121 & 0.0161 & 0.0105 &	0.0183 & 0.0161 & 0.0186 & 0.0195 \\
                      & 32 bits & 0.0254 & 0.0128 & 0.0156 & 0.0108 &	0.0224 & 0.0155 & 0.0219 & 0.0208 \\
                      & 64 bits & 0.0266 & 0.0167 & 0.0157 & 0.0127 & 0.0243 & 0.0146 & 0.0282 & 0.0239 \\
\hline
 \multirow{3}{*}{deep feature} & 16 bits & 0.4414 & 0.3327 & 0.4177 & 0.0148 & 0.3451 & 0.4375 & 0.5781 & 0.3638 \\
                      & 24 bits & 0.5301 & 0.4472 & 0.5102 & 0.0287 & 0.4359 & 0.5224 & 0.6045 & 0.4623 \\
                      & 32 bits & 0.5655 & 0.5001 & 0.5501 & 0.0351 & 0.4906 & 0.5576 & 0.6133 & 0.5168 \\
                      & 64 bits& \textbf{0.6148} & \textbf{0.5801} & \textbf{0.5956} & \textbf{0.0605} & \textbf{0.5718} & \textbf{0.6056} & \textbf{0.6273} & \textbf{0.5954} \\
\hline
\end{tabular}}
\end{center}
\caption{Comparison with shallow hashing competitors on QuickDraw-3.8M retrieval gallery.}
\label{table:shallow_hash_method_results}
\end{table*}

\begin{table}[tbp]
\vspace{0.3em}
\small
\begin{center}
\resizebox{\columnwidth}{!}{
\begin{tabular}{|p{1.5cm}<{\centering}|c|p{3.5cm}<{\centering}|}
\hline
 Splits & Number per category & Amount \\
\hline
\hline
 Training & 9000 & $9000\times345=3105000$ \\
 Validation & 1000 & $1000\times345=345000$ \\
 Retrieval & 1000 & $1000\times345=345000$ \\
 Query & 100 & $100\times345=34500$ \\
\hline
\end{tabular}}
\end{center}
\caption{Dataset splits on QuickDraw \cite{quickdraw2017} for our experiments.}
\vspace{-0.3cm}
\label{table:quickdrawdataset}
\end{table}

\section{Experiments}
\label{sec:experiments}
\subsection{Datasets and Settings}
\label{subsec: settings}
\noindent\textbf{Dataset Splits and Preprocessing}\quad Google QuickDraw dataset \cite{quickdraw2017} contains 345 object categories with more than 100,000 free-hand sketches for each category. Despite the large-scale sketches publicly available, we empirically find out that a number of around 10,000 sketches suffices for a sufficient representation of each category and thus randomly choose  9000, 1000 from which for training and validation, respectively. For evaluation, we form our query and retrieval gallery set by randomly choosing 100 and 1000 sketches from each category. A detailed illustration of the dataset split can be found at Table~\ref{table:quickdrawdataset}. Overall, this constitutes an experimental dataset of 3,829,500 sketches, standing itself on a million-scale analysis of sketch specific hashing problem, an order of magnitude larger than previous state-of-the-art research \cite{liu2017deepsketchhashing}, which we term as ``QuickDraw-3.8M".  We scale the raster pixel sketch to $224 \times 224 \times3$, with each brightness channel tiled equally, while processing the vector sketch same as with \cite{quickdraw2017}, with one critical exception -- rather than treating pen state as a sequence of three binary switches, \ie, continue ongoing stroke, start a new stroke and stop sketching, we reduce to two states by eliminating the sketch termination signal for faster training, leading each time-step input to the RNN module a four-dimensional input.


\noindent\textbf{Implementation Details}\quad Our RNN-based encoder uses bidirectional Gated Recurrent Units with two layers, with a hidden size of $512$ for each layer, and the CNN-based encoder follows the AlextNet \cite{krizhevsky2012imagenet} architecture with major difference at removing the local response normalization for faster training. We implement our model on one single Pascal TitanX GPU card, where for each pretraining stage, we train for $20, 5, 5$ epochs, taking about $20, 10, 10$ hours respectively. We set the importance weights $\lambda_{scl} = 0.01$ and $\lambda_{ql} = 0.0001$ during training and find this simple strategy works well. The model is trained end to end using the Adam optimizer \cite{kingma2014adam}. The learning rate starts at 0.01 and decays exponentially every 10 epochs by one order of magnitude. We report the mean average precision (MAP) and precision at top-rank 200 (precision@200), same with previous deep hashing methods~\cite{lin2015deep,zhao2015deepsemantic,liu2016deepsupervised,liu2017deepsketchhashing} for a fair comparison. Both the dataset and code will be made available from the SketchX website: \url{http://sketchx.eecs.qmul.ac.uk/downloads/}.

\subsection{Competitors}

We compare our deep sketch hashing model with several state-of-the-art deep hashing approaches and for a fair comparison, we evaluate all competitors under same base network if applicable. \noindent\textbf{DLBHC} \cite{lin2015deep} replaces our two-branch CNN-RNN module with a single-branch CNN module, with softmax cross entropy loss used for joint feature and hashing code learning. \textbf{DSH-Supervised}~\cite{liu2016deepsupervised} corresponds to a single-branch CNN model, but with noticeable difference in how to model the category-level discrimination, where pairwise contrastive loss is used based on the semantic pairing labels. We generate online image pairs within each training batch. \textbf{DSH-Sketch}~\cite{liu2017deepsketchhashing} is proposed to specifically target on modeling the sketch-photo cross-domain relations with a semi-heterogeneous network. To fit in our setting, we adopt the single-branch paradigm and their semantic factorization loss, where word vector is assumed to represent the visual category. We keep other settings the same.

We compare with six unsupervised (Principal Component Analysis Iterative Quantization (\textbf{PCA-ITQ)}~\cite{gong2013iterative}, Locality-Sensitive Hashing (\textbf{LSH})~\cite{andoni2006near}, Spectral Hashing (\textbf{SH})~\cite{weiss2009spectral}, Locality-Sensitive Hashing from Shift-Invariant Kernels (\textbf{SKLSH})~\cite{raginsky2009locality}, Density Sensitive Hashing (\textbf{DSH})~\cite{jin2014density}, Principal Component Analysis Hashing (\textbf{PCAH})~\cite{wang2006PCAH}) and two supervised (Supervised Discrete Hashing~(\textbf{SDH})~\cite{shen2015supervised}, Canonical Correlation Analysis Iterative Quantization (\textbf{CCA-ITQ})~\cite{gong2013iterative}) shallow hashing methods, where deep features are fed into directly for learning. It's noteworthy that running each of the above eight tasks needs about $100-200$ GB memory. Limited by this, we train a smaller model and use $256d$ deep feature (extracted from the fusion layer) as inputs.

\begin{figure}[thb]
	\centering
	\subfigure[\scriptsize{Precision-Recall curves @$16$ bits}]{
		\label{fig:ps_16bits}
		\includegraphics[width=0.23\textwidth]{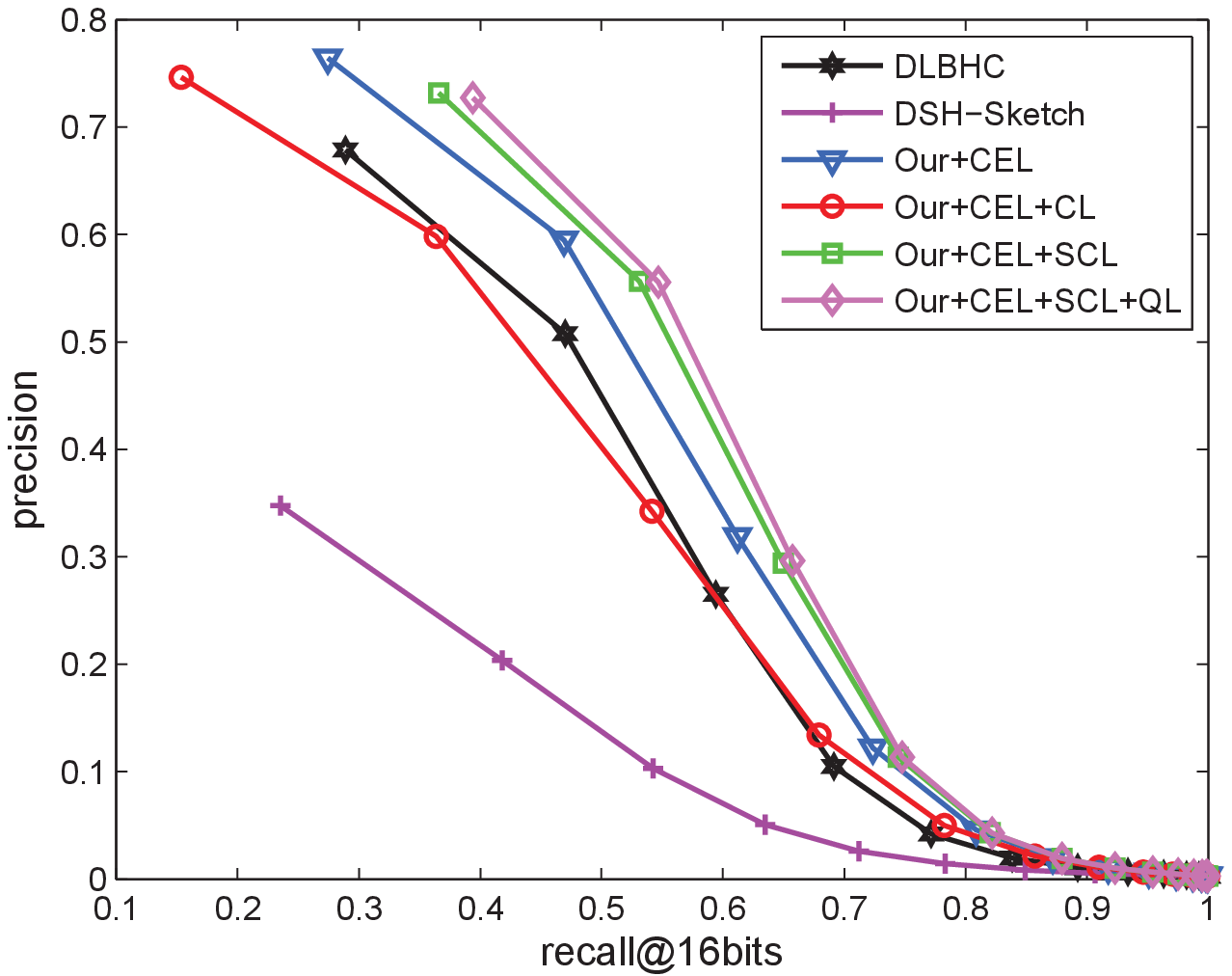}}
	\subfigure[\scriptsize{Precision-Recall curves @$24$ bits}]{
		\label{fig:ps_24bits}
		\includegraphics[width=0.23\textwidth]{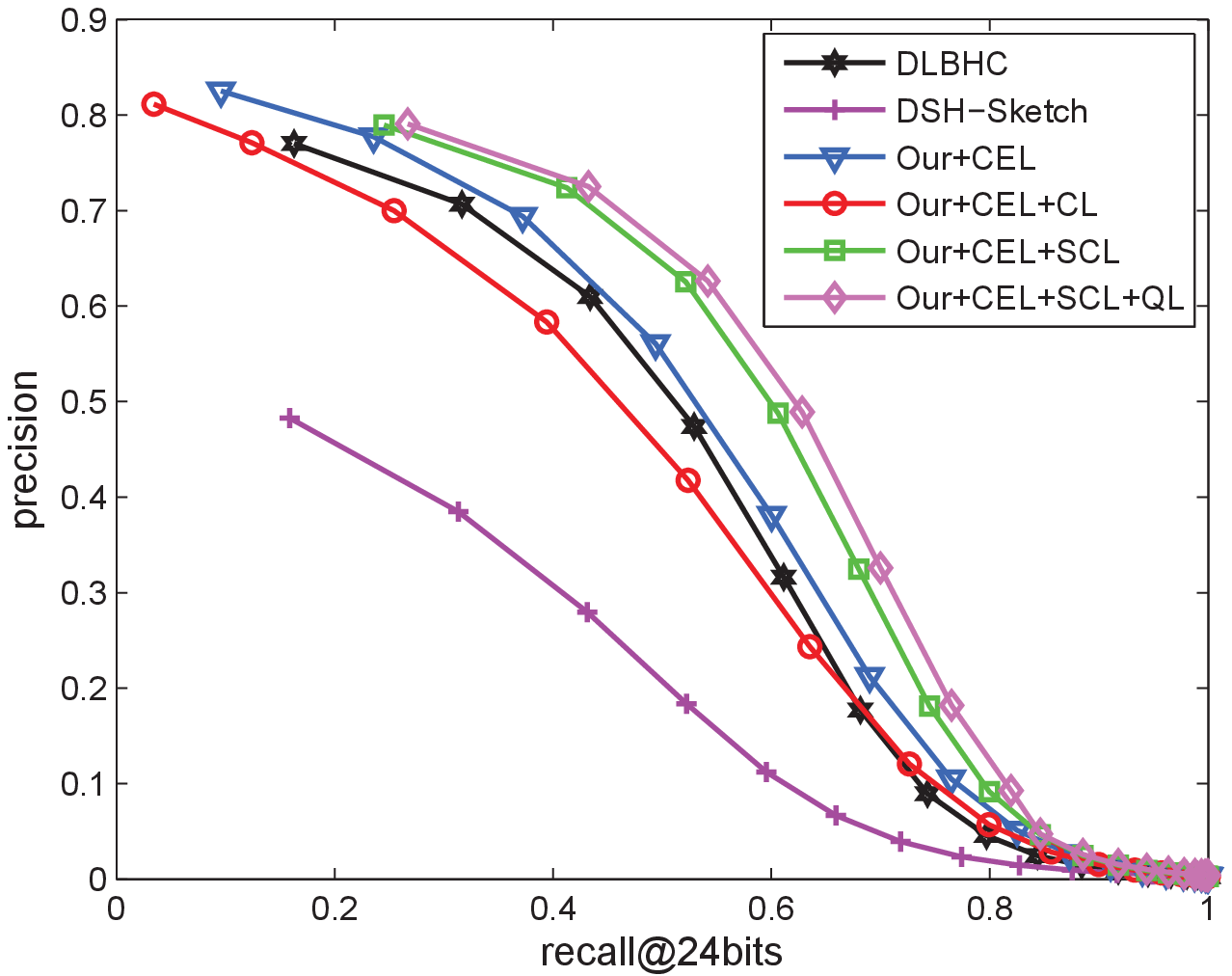}}
    \subfigure[\scriptsize{Precision-Recall curves @$32$ bits}]{
		\label{fig:ps_32bits}
		\includegraphics[width=0.23\textwidth]{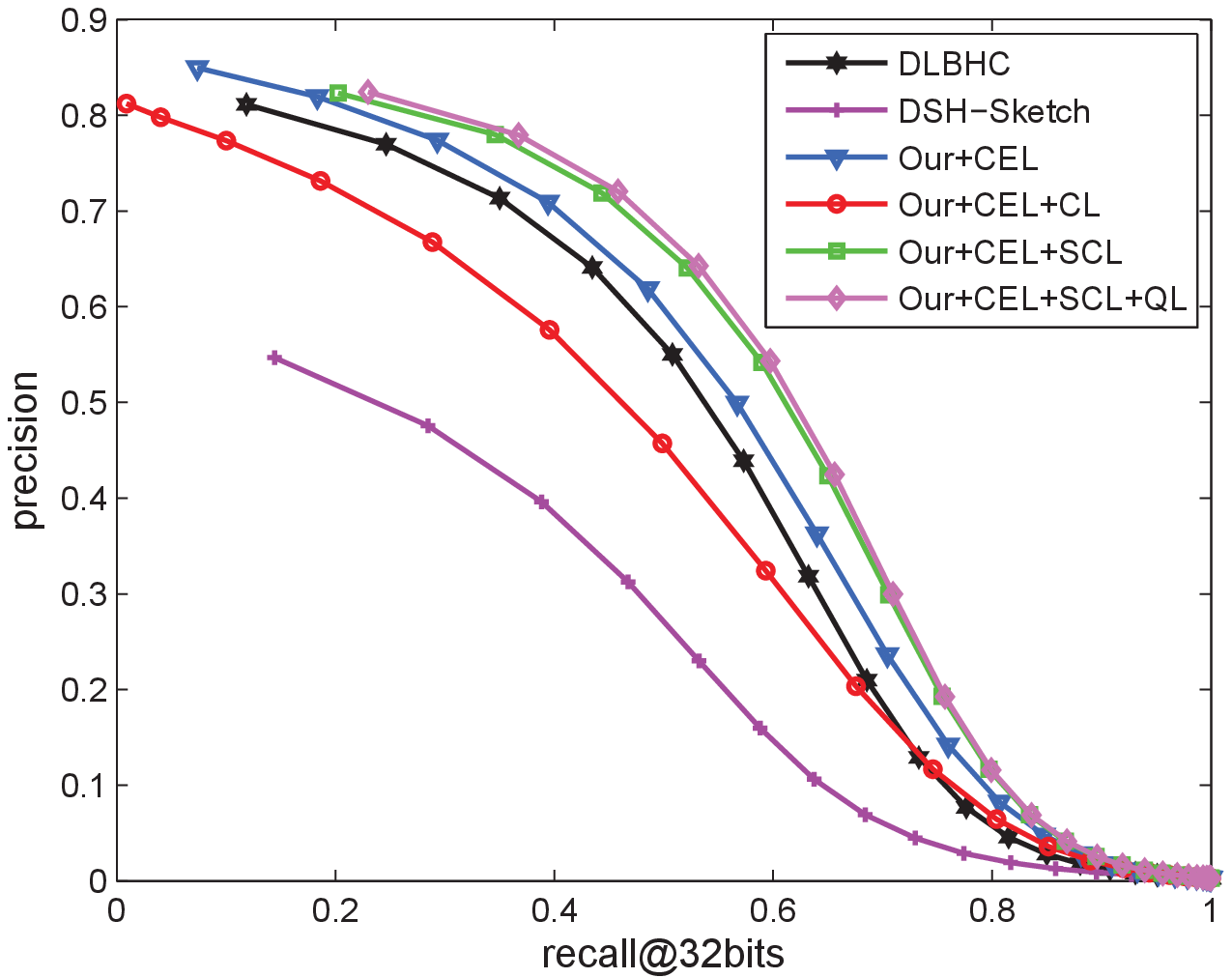}}
    \subfigure[\scriptsize{Precision-Recall curves @$64$ bits}]{
		\label{fig:ps_64bits}
		\includegraphics[width=0.23\textwidth]{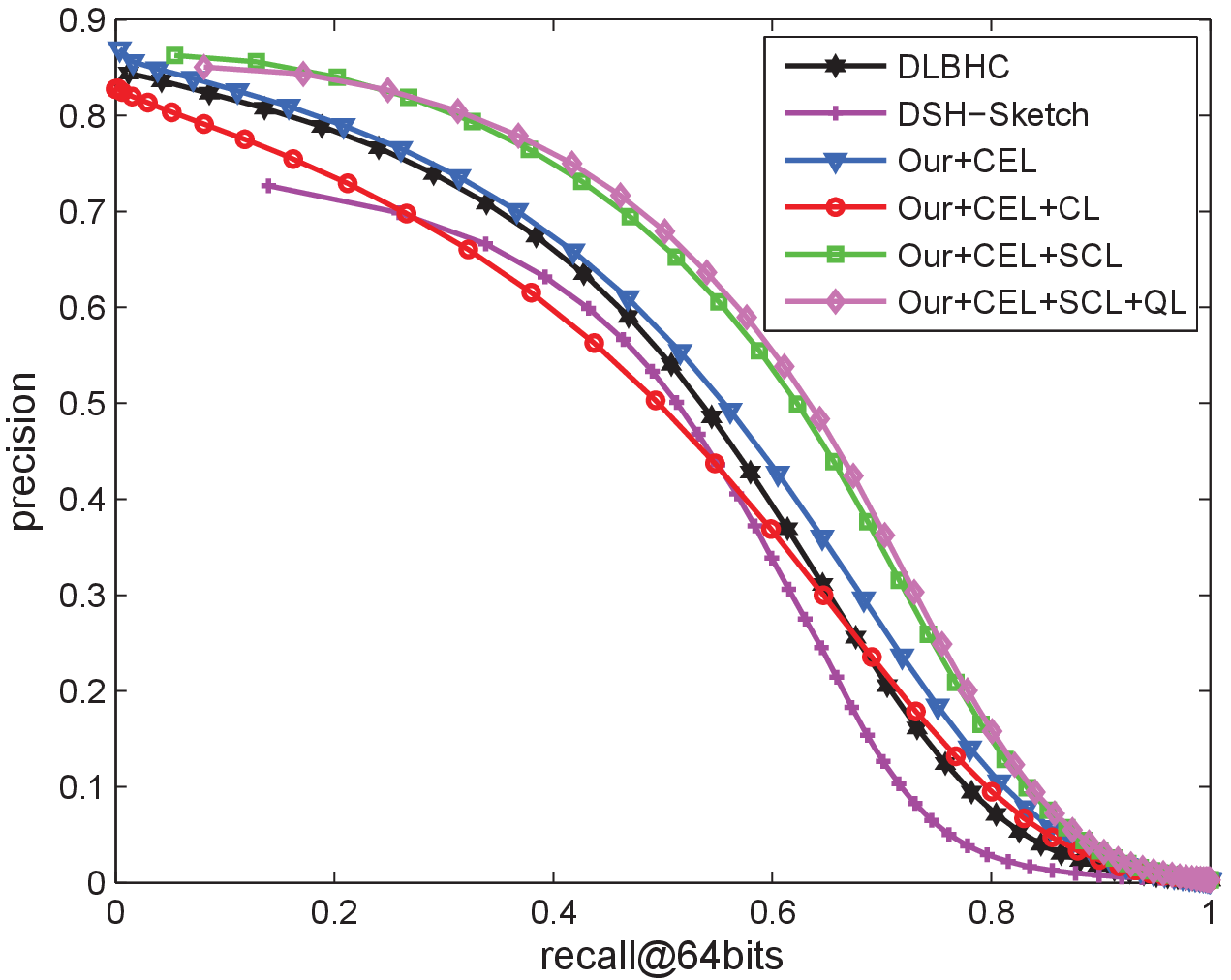}}
	\caption{Precision recall curves on QuickDraw-3.8M retrieval gallery. Best viewed in color.}
	\label{fig:precision_recall_curves}
\end{figure}

\subsection{Results and Discussions}

\noindent\textbf{Comparison against Deep Hashing Competitors} We compare our full model and the three state-of-the-art deep hashing methods. Table \ref{table:compare_with_deep_baseline} shows the results for sketch hashing retrieval under both metrics. We make the following observations: (i) Our model consistently outperforms previous state-of-the-art deep hashing methods by a significant margin, with 6.11/8.36 and 5.50/4.79 percent improvements (MAP/Precision@200) over the best performing competitor at 16-bit and 64-bit respectively. (ii) The gap between our model and DLBHC suggests the benefits of combining segment-level temporal information exhibited in a vector sketch with static pixel visual cues, the basis forming our CNN-RNN two-branch network, which may credit to (1) despite human tends to draw abstractly, they do share certain category-level coherent drawing styles, \ie, starting with a circle when sketching a sun, such that introducing additional discriminative power; (2) CNN suffers from sparse pixel image input \cite{yusketch2015BMVC} but prevails at building conceptual hierarchy \cite{mahendran2016visualizing}, where RNN-based vector input brings the complements. (iii) DSH-Supervised is unsuitable for retrieval across a large number of categories due to the incident imbalanced input of positive and negative pairs~\cite{lin2017discriminative}.
This shows the importance of metric selection under universal (hundreds of categories) million-scale sketch hashing retrieval, where softmax cross entropy loss generally works better, while pairwise contrastive loss hardly constrains the feature representation space and word vector can be misleading, \ie, basketball and apple are similar in terms of shape abstraction, but pushing further away under semantic distance.

\noindent\textbf{Comparison against Shallow Hashing Competitors} In Table \ref{table:shallow_hash_method_results}, we report the performance on several shallow hashing competitors, as a direct comparison with the deep hashing methods in Table \ref{table:compare_with_deep_baseline}, where we can observe that (i) shallow hashing learning generally fails to compete with joint end-to-end deep learning, where supervised shallow methods outperform unsupervised competitors; (ii) Under the shallow hashing learning context, deep features outperform shallow hand crafted features by one order of magnitude.

\begin{table}[!tbp]
\footnotesize
\centering
\resizebox{\columnwidth}{!}{
\begin{tabular}{|c|c|c|c|c|c|}
\hline
\multicolumn{2}{|c|}{\diagbox[dir=SE]{len}{dist}{model}}  & Our+CEL & Our+CEL+CL & Our+CEL+SCL & Our+CEL+SCL+QL (Full) \\
\hline \hline
\multirow{4}{*}{16 bits} & $d_1$             & 0.7501 & 0.5297 & 0.5078 & 0.5800 \\
                       & $d_2$             & 4.9764 & 3.2841 & 4.2581 & 4.8537 \\
                       & $d_1 / d_2$       & 0.1665 & 0.1721 & $\bm{0.1257}$ & 0.1290 \\
                       & MAP               & 0.5969 & 0.5567 & 0.6016 & $\bm{0.6064}$\\
\hline\hline
\multirow{4}{*}{24 bits} & $d_1$             & 1.2360 & 0.8285 & 0.6801 & 0.8568 \\
                       & $d_2$             & 6.1266 & 4.0388 & 5.0221 & 6.2243\\
                       & $d_1 / d_2$       & 0.2017 & 0.2051 & $\bm{0.1354}$ & 0.1377\\
                       & MAP               & 0.6196 & 0.5856 & 0.6374 & $\bm{0.6388}$\\
\hline\hline
\multirow{4}{*}{32 bits} & $d_1$             & 2.0066 & 1.5124 & 1.0792 & 1.2468 \\
                       & $d_2$             & 8.9190 & 7.3120 & 7.5340 & 8.6675 \\
                       & $d_1 / d_2$       & 0.2250 & 0.2068 & $\bm{0.1432}$ & 0.1439 \\
                       & MAP               & 0.6412 & 0.5911 & 0.6473 & $\bm{0.6521}$ \\
\hline\hline
\multirow{4}{*}{64 bits} & $d_1$             & 4.7040 & 3.5828 & 1.6109 & 2.5231\\
                       & $d_2$             & 15.4719 & 14.1112 & 11.6815 & 17.6179\\
                       & $d_1 / d_2$       & 0.3040 & 0.2539 & $\bm{0.1379}$ & 0.1432\\
                       & MAP               & 0.6525 & 0.6136 & 0.6767 & $\bm{0.6791}$ \\
\hline
\end{tabular}}
\vspace{0.1cm}
\caption{Statistic analysis for distances in the feature space of QuickDraw-3.8M under our model variants. $d_1$ and $d_2$ denote intra-class distance and inter-class distance, respectively.}
\label{table:distance_statistic_analysis}

\begin{center}
\resizebox{\columnwidth}{!}{
\begin{tabular}{|c|c|c|c|c|}
\hline
   &16 bit &24 bit &32 bit & 64 bit   \\
\hline
\hline
Retrieval time per query (s) & 0.089 & 0.126 & 0.157 & 0.286 \\
\hline
Memory load (MB) 345,000 gallery sketches & 612 & 667 & 732 & 937 \\
\hline
\end{tabular}}
\end{center}
\caption{Retrieval time (s) per query and memory load (MB) on QuickDraw-3.8M retrieval gallery.}
\label{table:running}
\end{table}

\noindent\textbf{Component Analysis} We have evaluated the effectiveness of different components of our model in Table \ref{table:compare_with_deep_baseline}. Specifically, we construct our model training with different loss combinations, including softmax cross entropy loss (Our+CEL), softmax cross entropy plus common center loss (Our+CEL+CL), softmax cross entropy plus sketch center loss (Our+CEL+SCL), softmax cross entropy plus sketch center loss plus quantization loss (Our+CEL+SCL+QL), which arrives our full model. We find that with cross entropy loss alone under our two-branch CNN-RNN model suffices to outperform best competitor, where by adding sketch center loss and quantization loss further boost the performance. It's noteworthy that adding common center loss harms the performance quite significantly, validating our sketch-specific center loss design. In Figure \ref{fig:precision_recall_curves}, we plot the precision-recall curves for all above-mentioned methods under $16$, $24$, $32$ and $64$ bit hashing codes respectively, which further matched our hypothesis.

\begin{figure*}[t]
\begin{center}
\includegraphics[width=0.85\textwidth]{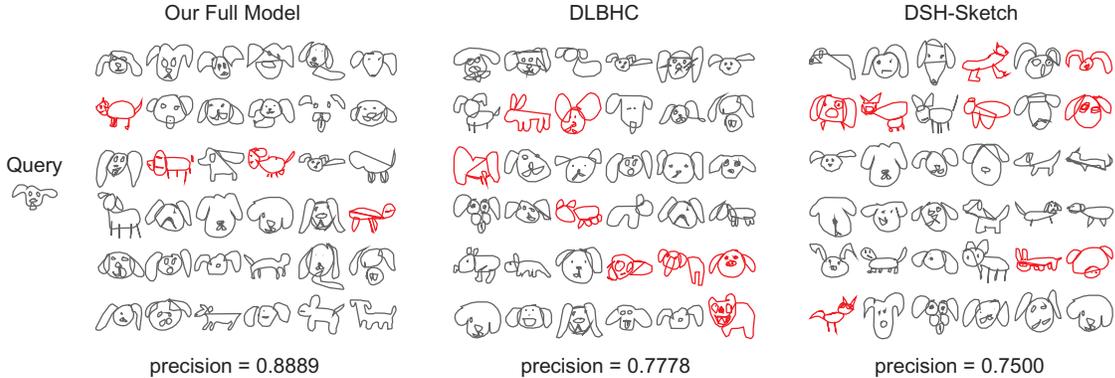}
\end{center}
   \caption{Qualitative comparison of top $36$ retrieval results of our model and state-of-the-art deep hashing methods for query (dog) at $64$ bits on QuickDraw-3.8M retrieval gallery. Red sketches indicates false positive sketch. The retrieval precision is obtained by computing the proportion of true positive sketch. Best viewed in color.}
\label{fig:qualitative}
\end{figure*}

\begin{table*}[!htbp]
\footnotesize
\begin{center}
\resizebox{\textwidth}{!}{
\begin{tabular}{|c|c|c|c|c|c|c|}
\hline
Model    & Sketch-a-Net~\cite{yu2017sketch} & ResNet $50$ ~\cite{he2016deep} & our RNN branch & our CNN branch & our RNN\&CNN + CEL & our RNN\&CNN + CEL + SCL  \\
\hline\hline
Accuracy & 0.6871 & 0.7864 & 0.7788 & 0.7376 & 0.7949 & \textbf{0.8051} \\
\hline
\end{tabular}}
\end{center}
\caption{Comparison with state-of-the-art methods and our model variants on sketch recognition task on QuickDraw-3.8M retrieval gallery.}
\label{table:classification_accuracies}
\end{table*}

\noindent\textbf{Further Analysis on Sketch Center Loss} To statistically illustrate the effectiveness of our sketch center loss, we calculate the average ratio of the intra-class distance $d_1$ and inter-class distance $d_2$, termed as $d_1/d_2$, among our $345$ training categories. A lower value of such score indicates a better feature space learning, since the objects within the same category tend to cluster tighter and push further away with those of different semantic labels, as forming a more discriminative feature space. In Table \ref{table:distance_statistic_analysis}, we witness significant improvement on the category structures brought by the sketch center loss across all hashing length setting (Our+CEL vs. Our+CEL+SCL), where on contrary, common center even undermines the performance (Our+CEL vs. Our+CEL+CL), which in accordance with what we've observed in Table \ref{table:compare_with_deep_baseline}.

\noindent\textbf{Qualitative Evaluation} In Figure \ref{fig:qualitative}, we qualitatively compare our full model with DLBHC \cite{lin2015deep} and DSH-Sketch \cite{liu2017deepsketchhashing} on the dog category. It's interesting to observe (i) how our model makes less semantic mistakes; (ii) how our mistake is more reasonably understandable, \ie, sketch is confusing in itself in most of our falsely-retrieved sketches, while in other methods some clear semantic errors take place (\eg, pigs and rabbits).

\noindent\textbf{Running Cost} We report the running cost as retrieval time (s) per query and memory load (MB) on QuickDraw-3.8M retrieval gallery (345,000 sketches) in Table \ref{table:running}, which even on million-scale can still achieve real-time retrieval performance.
\subsection{Generalization to Sketch Recognition}

To validate the generality of our sketch-specific design, we apply our two-branch CNN-RNN network to sketch recognition task, by directly adding a $2048d$ fully connected layer after joint fusion layer and before the 345-way classification layer. We compare with two state-of-the-art classification networks -- Sketch-a-net \cite{yu2017sketch} and ResNet-50 \cite{he2016deep}, where all above experiments are evaluated on the QuickDraw-3.8M retrieval gallery set. We demonstrate the results in Table \ref{table:classification_accuracies}, where following conclusion can be drawn: (i) Exploiting the sketching temporal orders is important, and by combining the traditional static pixel representation, more discriminative power is obtained ($79.49\% vs. 68.71\%$). (ii) Sketch center loss generalizes to sketch recognition task, bringing additional benefits.
\subsection{\textcolor{black}{Generalization to Zero-Shot Sketch Hashing}}

\textcolor{black}{We randomly pick $20$ categories from QuickDraw-3.8M and exclude them from training. We follow the same experimental procedures on $32$bit hash codes and report the MAP performance on the unseen categories. Under such challenging seen-unseen split, our method's MAP  of $0.7547$ outperforms that of DLBHC ($0.7094$) and DSH-Sketch ($0.5334$), by a clear margin.}
\section{Conclusion}
\label{sec:conclusion}
In this paper, we set out to study the novel problem of sketch hashing retrieval. By leveraging on a large-scale dataset of 3.8M human sketches, we explore the unique traits of sketches that were otherwise understudied in prior art. In particular, we show the benefit of stroke ordering information by encoding it in a CNN-RNN architecture, and we introduce a novel hashing loss that accommodates the abstract nature of sketches. Our hashing model outperforms all shallow and deep alternatives, and \textcolor{black}{yields superior generalization performance under a zero-shot setting and when re-purposed for sketch recognition.}

\textbf{Acknowledgment} This work was partly supported by NSFC No.$61773071$, BNSF No.$4162044$, Beijing Nova Program No. Z$171100001117049$, Z$181100006218137$, and BUPT Excellent PhD Student Foundation CX$2017307$.

{\small
\bibliographystyle{ieee}
\bibliography{egbib}
}

\end{document}